\documentclass[11pt]{article}

\usepackage[final]{acl}

\usepackage{times}
\usepackage{latexsym}

\usepackage[T1]{fontenc}

\usepackage[utf8]{inputenc}

\usepackage{microtype}

\usepackage{inconsolata}

\usepackage{graphicx}

\usepackage{amsmath}
\usepackage{placeins}

\title{Overcoming Copyright Barriers in Corpus Distribution Through Non-Reversible Hashing}

\author{
  \textbf{Arthur Amalvy\textsuperscript{1}},
  \textbf{Vincent Labatut\textsuperscript{2}},
  \textbf{Xavier Bost\textsuperscript{3}},
  \textbf{Hen-Hsen Huang\textsuperscript{1}}
  \\
  \textsuperscript{1} Institute of Information Science, Academia Sinica, Taiwan \\
  \texttt{arthuramalvy@as.edu.tw}, \texttt{hhhuang@iis.sinica.edu.tw} \\
  \textsuperscript{2} Laboratoire Informatique d'Avignon, Avignon Université, France \\
  \texttt{vincent.labatut@univ-avignon.fr} \\
  \textsuperscript{3} Aiway, France \\
  \texttt{xavier.bost@aiway.fr} \\ 
}

\begin{document}

\maketitle

\begin{abstract}
While annotated corpora are crucial in the field of natural language processing (NLP), those containing copyrighted material are difficult to exchange among researchers. Yet, such corpora are necessary to fully represent the diversity of data found in the wild in the context of NLP tasks. We tackle this issue by proposing a method to lawfully and publicly share the annotations of copyrighted literary texts. The corpus creator shares the annotations in clear, along with a \textit{non-reversible hashed} version of the source material. The corpus user must own the source material, and apply the same hash function to their own tokens, in order to match them to the shared annotations. Crucially, our method is robust to reasonable divergences in the version of the copyrighted data owned by the user. As an illustration, we present alignment experiments on different editions of novels. Our results show that our method is able to correctly align 98.7 to 99.79\% of tokens depending on the novel, provided the user version is sufficiently close to the corpus creator's version. We publicly release \texttt{novelshare}, a Python implementation of our method.
\end{abstract}


\section{Introduction}

Corpora, and in particular annotated corpora, are one of the central resources in the modern natural language processing (NLP) landscape. For many computational tasks, they are critical to train models, evaluate them and compare them against each other. Yet, while their importance is clearly established, researchers are often restricted to using freely available data released under a permissive license or in the public domain. This not only limits the amount of data available to the NLP community, but also introduces a systematic bias in studies since the performance of systems on copyrighted works is often not considered. For example, in the case of literary texts, recent works are protected, meaning most researchers restrict themselves to classical 19th plays and novels that fall in the public domain~\citep{Bamman2019-litbank,Han2021b}, with possible generalization issues~\citep{Lazaridou2021, Beelen2022}.

Ideally, a corpus creator should be able to freely share annotations to conform to open and reproducible science standards. It is important that data are made available online in a publicly accessible way, as on-demand sharing can be unreliable~\citep{Hussey2025}. In the specific case of copyrighted data, sharing should only be possible if the user is also in possession of the original material. A naive strategy could be to share the corpus annotations along with instructions to obtain the data, and ensure their data are exactly identical to the creator's. The user could then use their data jointly with the shared annotations, and copyright would be enforced. However, in practice, the user's data are rarely exactly identical to the creator's. Since copyrighted works are not always directly available online, the user may obtain a non-identical version that can differ in multiple ways, such as how the digitization process was carried out. The data may also differ because of how they were prepared to the format required for NLP experiments. Therefore, we need an annotation-sharing scheme that is robust enough to handle reasonable divergences in the user's material while providing copyright compliance guarantees. 

\begin{figure*}[htb]
    \centering
    \includegraphics[width=1\linewidth]{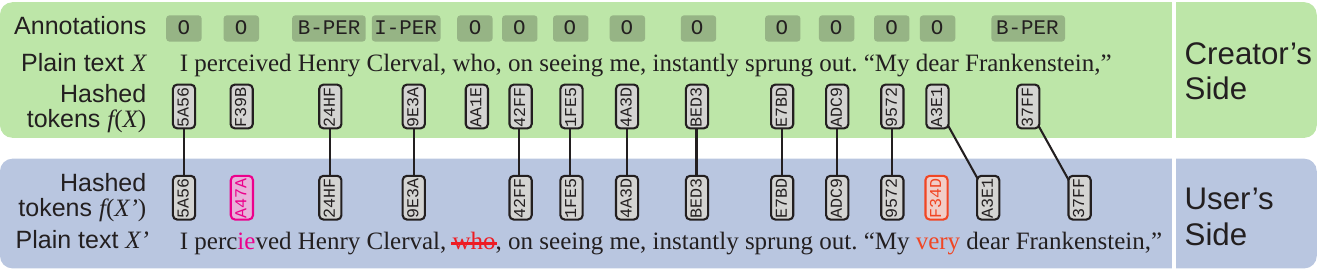}
    \caption{Proposed hashing scheme. Top part: the corpus creator's plain text ($X$) and hashed tokens $f(X)$, and corresponding NER annotations. Bottom part: the corpus user's plain text ($X'$) and hashed tokens ($f(X')$. Our method matches both sequences of hashes, while supporting small differences such as typos (shown in purple), missing tokens (red) and additional tokens (orange). The user never accesses the creator's plain text.}
    \label{fig:encr-scheme}
\end{figure*}

Motivated by this issue and inspired by a previous attempt by \citet{Bost2020-serial_speakers}, we propose a method to easily share a corpus under copyright constraints, in the case where the user possesses material that is reasonably close to the creator's. We place ourselves in the situation where the corpus to share is composed of a sequence of tokens, and one or more annotations for each token. The named entity recognition (NER) task is a good example of such a situation, where each token is annotated with a tag in the BIO scheme~\citep{Ramshaw1995-iob}. We hash each token of the creator's corpus with a non-reversible cryptographic function, and shorten each resulting code to voluntarily create collisions and avoid attacks based on precomputed hash tables. On the user's side, each token is hashed using the same method. We robustly match the creator's and user's truncated hashes, which allows us to align annotations with user's tokens. Since the two corpora may be marginally different, it is likely that some tokens remain unaligned during this first stage. Therefore, we propose additional strategies to align some of these remaining tokens. We present this overall process in Figure~\ref{fig:encr-scheme}.

To validate the effectiveness of our sharing technique, we carry multiple experiments on a corpus of three novels, each one coming in three different editions. We empirically show that our method can accurately align almost all the hashed tokens, even when the user owns a different edition from the creator. Furthermore, our experiments validate the interest of our additional alignment strategies, that can be used to increase the number of correctly aligned tokens.
 
We release all the source code and data needed to reproduce our experiments under a free license\footnote{\url{https://github.com/CompNet/novelshare}}. Additionally, we publicly release \textit{novelshare}, an implementation of our alignment method that can be used to share sequential annotations of copyrighted texts.

\section{Related Work}

\subsection{NLP Corpus Sharing}
\label{sec:RelWorkNlp}

The most popular contemporary large language models come from the industry and rely on extremely large collections of textual data drawn from a wide range of sources, including books, journalistic content, and other works protected by intellectual property rights~\cite{Henderson2023, Ahmed2026}. This shows the importance of such copyrighted material in tackling a variety of NLP tasks. Because these corpora are often assembled through large-scale Web crawling and aggregation, they frequently contain protected material for which no direct licensing or authorization has been obtained, prompting ongoing debates about the legal status of such practices and the lack of transparency surrounding training data composition~\cite{Buick2024}. Academic researchers typically do not have access to the same legal and financial resources as these large firms, and therefore adopt various strategies to work around copyright constraints. 

The first option is to constitute corpora based only on \textit{public domain} data. This is for example the case of the well-known literary corpus Litbank~\citep{Bamman2019-litbank} and its French equivalent fr-Litbank~\citep{Melanie-Becquet2024}, whose most recent novels are from 1922 and 1937 respectively. Similarly, the speaker attribution corpora QuoteLi3~\citep{Muzny2017-sieve_quote_attribution} and PDNC~\citep{Vishnubhotla2022-PDNC} focus only on classic texts, and the coreference resolution corpus FantasyCoref~\citep{Han2021b} only includes \textit{Alice in Wonderland} and public domain fairy tales. This reliance on older, public domain texts in such corpora is a fundamental problem when using them to train NLP models. Since these models are often used on contemporary texts, training on older data systematically biases model evaluation and hinders their performance~\cite{Lazaridou2021, Beelen2022}. Additionally, public domain texts are often part of the training data of large models, which further increase evaluation bias concerns due to data contamination~\citep{Johnson2024}. 

The second strategy to avoid legal problems is simply to not share at all any corpus that includes protected material. For example, \citet{Camp2012} create a corpus containing biographies annotated for the identification and classification of personal relationships, but do not share this copyrighted material. \citet{Chun2025} extract character networks from Korean dramas, but completely anonymize their data to the point of not providing even the titles of the considered works. Of course, this practice is not on par with modern NLP standard, as it hinders reproducibility and the ability of researchers to draw comparisons. 

The third approach to limit copyright infringement issues is to use only \textit{excerpts} of the original texts when building a corpus. The rationale here is that this practice, as it concerns academic research, could fall under fair use. For instance, \citet{Dekker2019-charnets} propose a corpus of contemporary novels annotated for NER, and explicitly state that they only share the first chapters of these novels because of copyright concerns. Under Chinese law, \citet{Zhao2025a} can use up to 10 chapters by novel to constitute their GenWebNovel NER corpus. This practice illustrates the legal ambiguity faced by academic researchers, as publicly sharing annotated excerpts of copyrighted literary works goes beyond what is clearly permitted under standard research exceptions in many jurisdictions. From the NLP perspective, working on full novels is necessary to tackle problems related to text length~\cite{Amalvy2024e, Bourgois2025}.

Fourth, it is lawfully possible to share full copyrighted texts, provided their authors agree. Some researchers therefore explicitly ask for permission. For instance, \citet{Johnson2024} obtain the consent of each individual author to create \textit{FicSim}, their corpus of fan-fictions. \citet{Alrashid2023} ask an author permission to include his kid story in ScANT, their corpus of scene-annotated narrative texts. While this is an ethically sound way of releasing data, it is unrealistic to rely on such a costly method, even to constitute relatively small corpora. 

Finally, the fifth solution to the issue of sharing annotated copyrighted material is proposed by \citet{Bost2020-serial_speakers}. In order to share the copyrighted TV series dialogues constituting their \textit{Serial Speakers} corpus, they design an hash-alignment scheme that allows any user possessing the original dialogues to access the annotations. This approach is the only one allowing to share annotations of full copyrighted texts in a lawful way, as it does not share the original text but a \textit{non-reversible hashed version}. We discuss this legal aspect further in Appendix~\ref{sec:Legal}. We expand on this method by studying the effects of its parameters, generalizing it to any sequence of annotated tokens, and proposing additional strategies to improve alignment performance.

\subsection{Related Problems}

While this issue of copyrighted corpus sharing is not explicitly identified in NLP except by \citet{Bost2020-serial_speakers}, it resembles other problems from different fields. \textit{Sequence reconstruction} is a class of problems introduced by~\citet{Levenshtein2001-seqa,Levenshtein2001-seqb} where the goal is to reconstruct a sequence from a set of partial and sometimes noisy observations~\citep{Wei2024}. The \textit{trace reconstruction problem}~\citep{Batu2004-trace_reconstruction} is also related: in that setup, a binary string is passed through a deletion channel that may delete each of its bits with a set probability $p$, resulting in a shorter string called a \textit{trace}. The goal is to determine how many such traces are needed to reconstruct the original sequence with a high probability. Both of these tasks differ from ours: the user does not have access to a hashed non-noisy version of the original sequence, both problems assume the multiplicity of observed distorted sequences and the noise observed in our setup is specific to our application. Similar problems also arise in the context of DNA sequencing, where genomic data consisting of DNA sequences must be communicated and used securely in public cloud environments~\citep{Mete2015-dna_privacy,Lu2021-dna_privacy}. For \textit{read mapping}, a fundamental operation consisting in aligning a set of short DNA sequences to a reference genome, an existing strategy consists in sharing only hashed sequences with non-trusted cloud environments to perform part of the computation~\citep{Chen2012-dna_privacy,Kang2016-dna_privacy}. This differs from our setting as we need to align freely shareable annotations to the user's version of the corpus.

\section{Methods}

\subsection{General Principle}
Let $X$ be a sequence $(x_1,\dotsc, x_n)$ of copyrighted tokens, annotated by a \textit{corpus creator}. Each token in $X$ can have one or more annotations, that correspond to task-specific labels. Such an annotation scheme encompasses token-level tasks such as token classification, but also supports coarser annotation at the span level or higher: for example, the BIO NER scheme~\citep{Ramshaw1995-iob} allows to annotate entity spans at the token level. These annotations are not copyrighted, and consist of one or more different sequences of the same length as $X$, that can be freely shared. The creator wants to share these annotations with a \textit{corpus user}, but without making the tokens $X$ public, as they are copyrighted. For this purpose, we hash them in a non-reversible way, producing a new sequence $f(X) = (f(x_1),\dotsc,f(x_n))$ which is shared along with the annotations. This is illustrated by the top part of Figure~\ref{fig:encr-scheme} (green block), where we consider a NER example. As can be seen in the figure, each token of the corpus has a tag attached that indicates whether or not this token is part of an entity.

On their side, the user must possess the tokens too, in order to match them to the annotations shared by the creator. However, it is very unlikely that the user has access to the \textit{exact} same sequence $X$ as the creator. In the case of novels, for instance, turning a text into a token sequence usable for NLP experiments is a multi-step process involving many parameters. 
First, the user might have a different version of the novel, including editorial differences such as corrections, revisions, or a modernized text. Second, the process of digitization necessary to obtain an electronic book relies on some technical choices that can vary depending on place, time, and publisher: punctuation and typographic conventions, decomposition in chapters, text encoding. This step may even require error-prone steps such as optical character recognition. Finally, extracting a token sequence from the electronic book also necessitates to make methodological choices that can differ from the creator's, especially regarding text tokenization. 

For all these reasons, it is reasonable to assume that the user possesses a slightly different token sequence, noted $X' = (x'_1,\dotsc,x'_m)$. In order to get the annotations associated to these tokens, the user must first align their tokens with the creator's. For this purpose, we use the same hashing method as the creator, to produce a new sequence $f(X') = (f(x'_1),\dotsc,f(x'_n))$. This part is represented in the bottom part of Figure~\ref{fig:encr-scheme} (blue block). The alignment is then performed only by comparing the hashes. 

At this stage, it is crucial to stress two essential methodological points that establish the lawfulness of our method as discussed in Section~\ref{sec:RelWorkNlp} and Appendix~\ref{sec:Legal}. First, \textbf{the creator's plain text is \textit{never} shared with the user}: only the hashed tokens and their corresponding annotations are. Second, \textbf{our method does not try to decrypt the tokens} to obtain creator's plain text: it works only with the user's plain tokens, which must therefore be as similar as possible to the creator's.

\subsection{Hashing}

To hash the copyrighted sequence $X$, we process each token using the SHA-256 cryptographic function, resulting in hashed sequence $f(X)$. We pick SHA-256 for its wide availability, usage and its known robustness to inversion attempts. In practice, hashing an entire novel is near-instantaneous.

The rate of collision (cases where the hash function computes the same output for different inputs) of SHA-256 is, by design, extremely low. Since an attacker can easily acquire a precomputed hash table with every possible word for a specific language, it would be trivial to break such a naive scheme. Therefore, we truncate each hash to forcibly increase the collision rate of the hash function. Thus, even if the attacker computes the hash of each possible word in the vocabulary, this only allows them to narrow down the set of possible tokens for a given hash to a set of words, but they have no way of knowing which of these words is the correct one.

On the user's side, we proceed similarly and apply the same hash function to the plain tokens $X'$ in order to produce hashed sequence $f(X')$.

\subsection{Naive Alignment}

The next step consists in applying any alignment algorithm between $f(X)$ and $f(X')$, through exact matching. As a consequence, when two hashes $f(x_i)$ and $f(x'_j)$ are matched, they are necessarily equal. In this case, we assume that $x_i = x'_j$, allowing us to determine which annotation is associated to $x'_i$. It is worth stressing that this assumption is not guaranteed to be correct, since we use truncated hashes that lead to collisions. In practice though, we find that such alignment method is sufficiently robust, since the context tokens disambiguate these collisions as we show in Section~\ref{sec:hash-len}.

\subsection{Additional Alignment Strategies}
\label{sec:Strat}

Due to the differences between $X$ and $X'$, it is likely that the above naive alignment method will not be able to align all the hashes. We identify three distinct situations:

\paragraph{Addition} The user provides superfluous hashes that do not correspond to any hashes in the creator's sequence. As an example, this is the case of the orange token \textit{``very''} in Figure~\ref{fig:encr-scheme}. 

\paragraph{Deletion} The user does not provide one or more hashes, like the missing token \textit{``who''} in Figure~\ref{fig:encr-scheme}.

\paragraph{Substitution} The user provides certain hashes that should be aligned with some of the creator's hashes, but their values are different. In Figure~\ref{fig:encr-scheme}, there is a typo in the token \textit{``perceived''} on the user's side, resulting in the token \textit{``percieved''} (in purple) causing the substitution.

\medskip

In the addition case, we can safely discard the superfluous user's tokens, as there are no corresponding annotations in the creator's sequence. The deletion and substitution cases are more challenging, which is why we handle them through additional strategies aimed at being applied \textit{after} the initial naive alignment.

\begin{figure}[htb]
    \centering
    \includegraphics[width=0.8\linewidth]{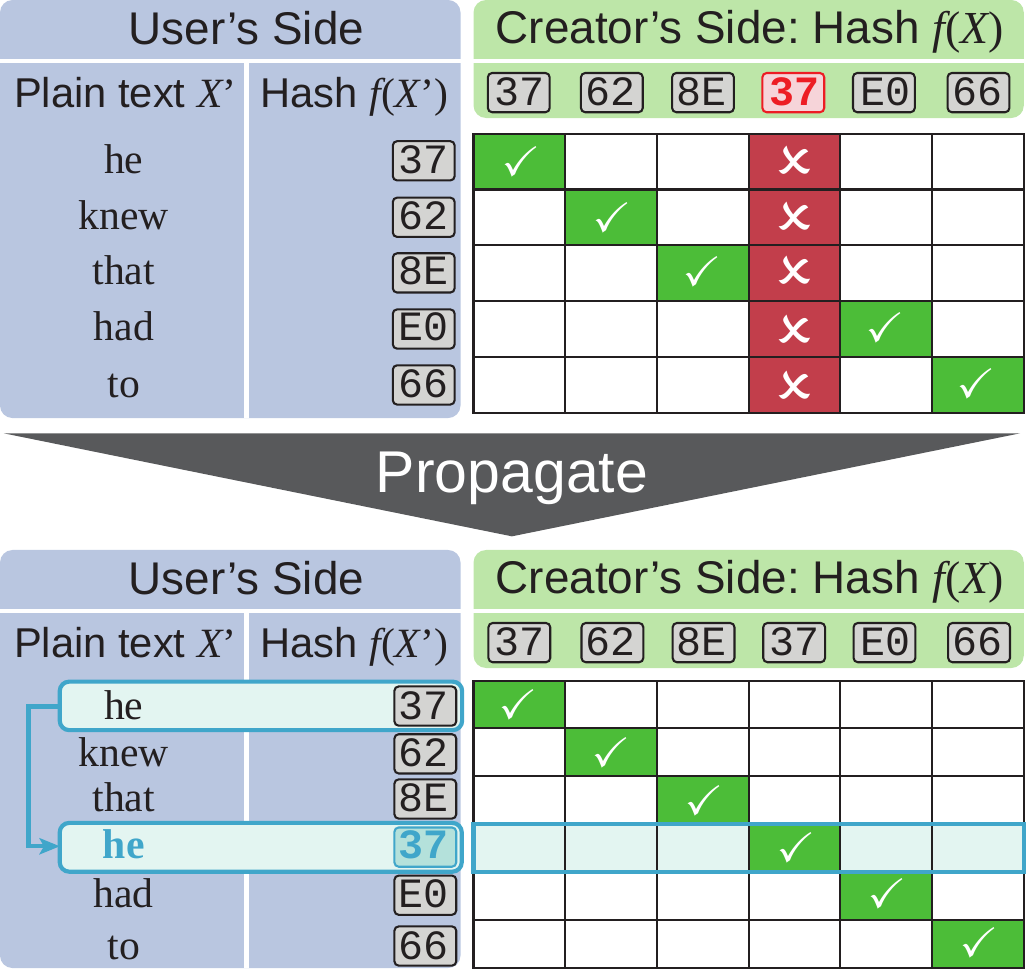}
    \caption{An example of applying the \texttt{propagate} strategy to retrieve a token missing on the user's side (shown in red), by leveraging the fact that the same hash value (\texttt{37}) is matched in another location (shown in cyan). Note that the creator's plain text $X$ is not available at the alignment stage.}
    \label{fig:propagate}
\end{figure}

\paragraph{\texttt{propagate} strategy} When a creator's hash $f(x_i)$ cannot be matched due to a missing or substituted user's hash, it is possible that other tokens $x_j$ with the same hash $f(x_j) = f(x_i)$ were already aligned with some user's hashes at different points in the sequence. In that case, we proceed to a vote and set the token at position $i$ as the majority token in the user's sequence. We thereby "propagate" decisions made at the previous stage to hashes still pending alignment. Figure~\ref{fig:propagate} shows an example of applying the \texttt{propagate} strategy on a deletion.

\begin{figure}[htb]
    \centering
    \includegraphics[width=0.8\linewidth]{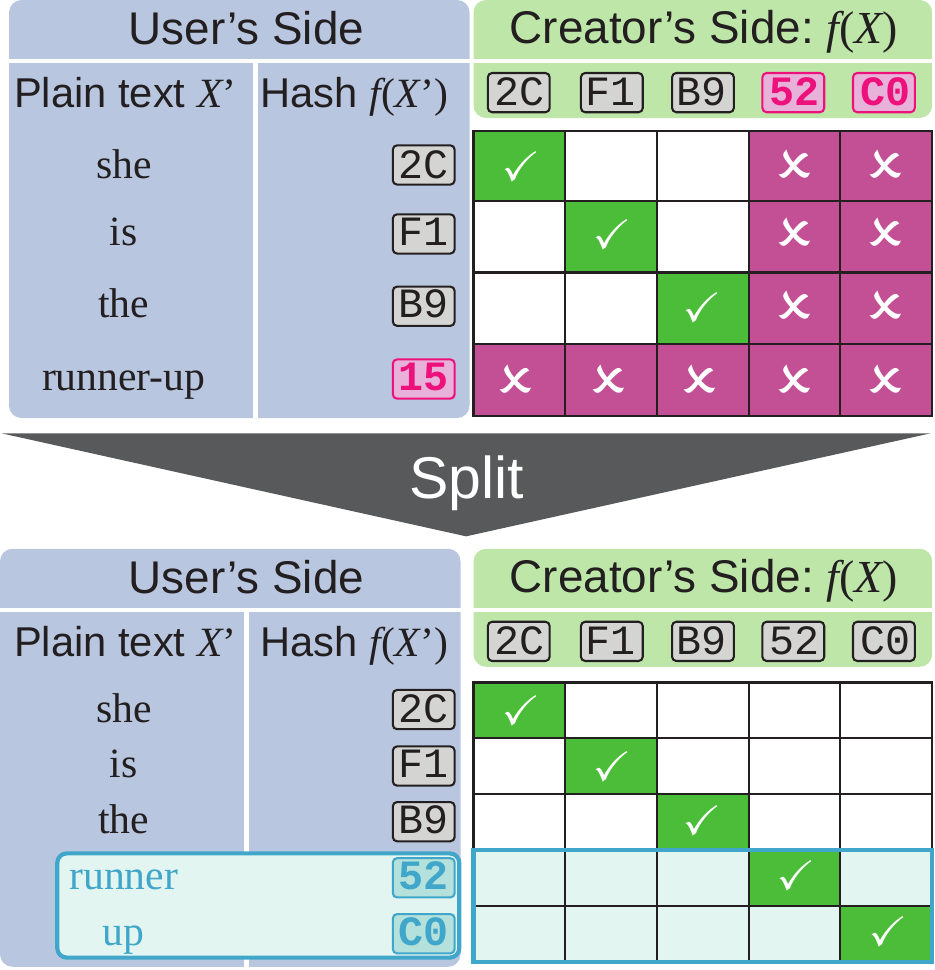}
    \caption{An example of applying the \texttt{retokenize} strategy on a substitution case (shown in purple). The corpus creator and user perform a different tokenization, resulting in tokens \textit{``runner''} and \textit{``up''} vs. \textit{``runner-up''}, respectively. By testing all possible splits of the user's token, the strategy is able to recover from this error (in cyan).}
    \label{fig:split}
\end{figure}

\paragraph{\texttt{retokenize} strategy} In the case of a substitution, there is a possibility that the creator's tokens underwent a different tokenization compared to the user's. Figure~\ref{fig:split} shows such an example, where the creator's tokens \textit{``runner''} and \textit{``up''} correspond to the user's token \textit{``runner-up''}. To resolve this case, we iterate through all possible splits of the user's token, hash them and compare them to the creator's  hashes. If they match, we keep this split in the user's sequence $X'$. In the reverse case where the user's tokens have been incorrectly split, we merge them and compare the subsequent hash to the aligned creator's hash. 

\paragraph{\texttt{case} strategy} Certain substitutions are due to a different casing between the creator's and user's tokens $x_i$ and $x'_j$. To handle this situation, we try different casing options $c()$ for $x'_j$ and recompute its hash. If $f(c(x'_j))$ matches $f(x_i)$, we align $x_i$ with $c(x'_j)$ in the user's sequence.

\paragraph{\texttt{mlm} strategy} When a token is missing in the user's sequence (either because of a deletion or a substitution), we leverage the textual context available on the user's side and try to estimate this token using masked language modeling (MLM). We insert a \texttt{[MASK]} token at the concerned position in $X'$, and use a pretrained model to predict the most likely word. If its hash matches the creator's hash $f(x_i)$, we keep the word in $X'$.

\paragraph{\texttt{pipe} meta-strategy}  We combine the above strategies within the \texttt{pipe} meta-strategy, where we sequentially apply them in a predefined order to try and fix multiple classes of sequence differences. 

\smallskip
The above strategies are language-agnostic, except for \texttt{case}, which only makes sense for scripts with a concept of letter case, such as the Latin script; and \texttt{mlm}, which necessitates to have a trained model that supports the language of interest. 
It is important to stress that these strategies do not aim at aligning the user's text at any cost. If this text is too different from the creator's, the method \textit{should} fail: trying to reconstruct the creator's text with MLM would go against international copyright laws. Our proposed strategies implement a tradeoff between, on the one hand, robustness to minor differences in the user's text, and, on the other hand, being copyright-compliant.

\section{Experiments}

We now perform alignment experiments on real novels to validate the effectiveness of our method. 

\subsection{Corpus}

Our corpus is based on three public domain novels: Mary Shelley's \textit{Frankenstein}, Herman Melville's \textit{Moby Dick} and Jane Austen's \textit{Pride and Prejudice}. For each one, we gather three editions. We consider the earliest one as the \textit{creator's edition}, used to produce the annotated token sequence of the corpus creator, while both remaining editions are alternative \textit{user's editions}. Among the latter, one is commonly considered as \textit{close} to the creator's edition, whereas the other is more \textit{distant}. We hypothesize that a closer edition should allow for better alignment performance. We provide more information on the exact sources we use for each edition in Appendix~\ref{apdx:data-details}.

\paragraph{\textit{Frankenstein}} The first edition was first published in 1818 (\texttt{F-1818}, creator's). A later 1823 edition (\texttt{F-1823}, close) came with minor changes, and remains close to the original. Meanwhile, the 1831 edition (\texttt{F-1831}, distant) is a version of the novel revised by its author, with many significant differences: for example, the original first chapter was expanded and split in two.

\paragraph{\textit{Moby Dick}} This novel was originally published in 1851 both in the USA (\texttt{MD-1851-US}, creator's) and the UK (\texttt{MD-1851-UK}, distant). These editions differ significantly from one another though, as the UK edition was censored and modified heavily and independently by its editor, which led for example to the removal of Chapter~25. Additionally, although the reason for that change is unknown, the epilogue is also missing in this version, changing the end of story. Finally, we also include the 1988 Northwestern-Newberry edition (\textit{MD-1988}, close), which is closer to the original US edition.

\paragraph{\textit{Pride and Prejudice}} By contrast, this novel had a simpler editorial life, its later editions only differing in small changes such as modernized spelling. The first edition came in 1813 (\texttt{PP-1813}, creator's), and the second one in 1817 (\texttt{PP-1817}, close). We also include the later 1894 illustrated edition (\texttt{PP-1894}, distant).

\medskip

In addition to these novels, we also experiment on other text domains (web and news) in Appendix~\ref{apdx:results-other-domains}.

\subsection{Setup}
\label{sec:xp-details}

In all of our experiments, we use the enhanced version of the gestalt pattern matching alignment algorithm~\citep{Ratcliff1988-diff} implemented by the \texttt{difflib} module in the Python standard library. Since the algorithm is quadratic in time for the worst case scenario, we optimize its runtime by aligning novels in our corpus chapter by chapter. We deem this optimization acceptable as it is unlikely that tokens should be aligned across chapters. We only apply this optimization when the number of chapters is the same between the creator's and user's novels, otherwise we align directly on the entire content (\texttt{F-1831}, \texttt{MD-1851-UK}). 

We use the \texttt{ModernBERT-base} model~\citep{Warner2025-modernbert} for the \texttt{mlm} strategy, with a window size of 32 (see Appendix~\ref{apdx:mlm-window} for details). For the \texttt{pipe}  meta-strategy, we use the sequence \texttt{retokenize}, \texttt{mlm}, \texttt{case}, \texttt{propagate}. This order prioritizes high-precision strategies (see Appendix~\ref{apdx:pipe-order} for details).

\subsection{Effect of Truncated Hash Length}
\label{sec:hash-len}
We first study the effect of the length of our truncated hash on performance and security. For each novel, we hash the creator's, close and distant editions, and align the resulting hashes with our proposed method. Lowering the length of the hash creates more collisions, improving security but also increasing the risk of errors as aligning tokens is more difficult and some alignment strategies may be impacted.

\begin{figure}[htb]
    \centering
    \includegraphics[width=1.0\linewidth]{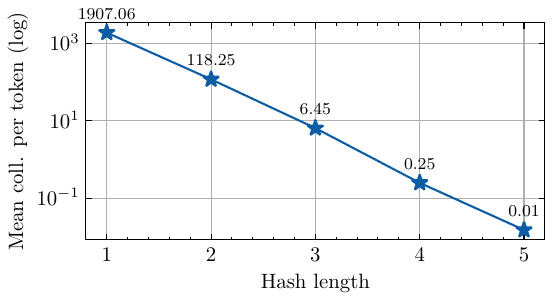}
    \caption{Mean number of hash collisions per token in our experimental corpus. We exclude hash lengths higher than 5 since they are too close to 0.}
    \label{fig:hash_collisions}
\end{figure}

Figure~\ref{fig:hash_collisions} shows the mean number of collisions per token in our corpus depending on hash length. Given this figure, values 1 ($1907.06$ collisions per token), 2 ($118.25$) or 3 ($6.45$) appear as appropriate candidates. In these cases, on average, a potential attacker has to choose between multiple token possibilities, but has no way of confirming that their choice is correct without access to the original data. For longer hashes, the number of collisions is close to 0, which we deem not secure enough.

\begin{figure}[htb]
    \centering
    \includegraphics[width=\linewidth]{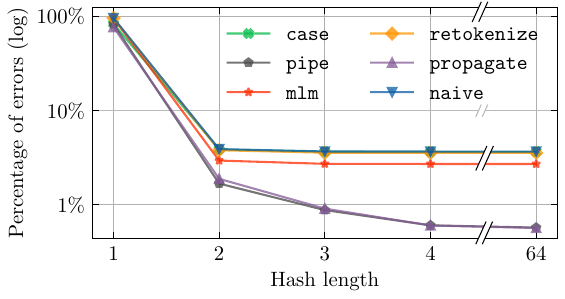}
    \caption{Mean percentage of errors across editions as a function of hash length (\{1, 2, 3, 4, 64\}) for different alignment strategies}
    \label{fig:hash_len}
\end{figure}

To choose the best hash length, we have to observe its impact on alignment performance. We do so by performing our experiment with hash lengths in $\{1, 2, 3, 4, 64\}$. Figure~\ref{fig:hash_len} shows the mean percentage of token alignment errors across editions. Decreasing the hash length increases the number of errors for all strategies, highlighting the necessary tradeoff between security and alignment reliability. Additionally, we also observe that some alignment strategies are more sensitive to hash length than others. The \texttt{propagate} strategy is particularly vulnerable to truncation, as it may propagate errors rather than correctly align tokens. On the other end of the spectrum, the \texttt{case} and \texttt{mlm} strategies are less sensitive. For the rest of the experiments, we present results with a hash length of 2 as a tradeoff between security and alignment performance.

\subsection{Results Per Edition}
\label{sec:ResEditions}
In the ideal case, we would expect the user to own exactly the same plain text as the creator, i.e. the same edition of the novel. However, here we consider a more difficult situation, where the user owns a different edition (close, distant) compared to the creator. We compare the impact of our proposed alignment strategies by recording the number of incorrectly aligned tokens, and plot our results in Figure~\ref{fig:errors-percent}. As a practical illustration, we also apply our alignment method to NER in Appendix~\ref{apdx:ner-results}, where it is able to align 96.48\% of entities.

\begin{figure}[htb]
    \centering
    \includegraphics[width=1.0\linewidth]{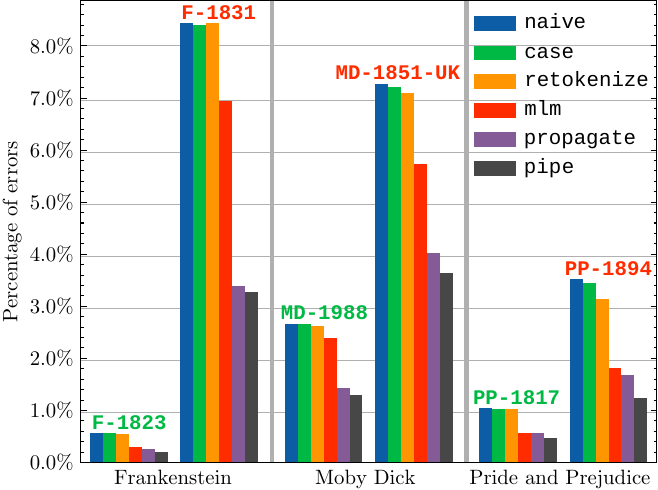}
    \caption{Percentage of misaligned tokens depending on the strategy and user's edition. For each novel, the user's edition which is the closest to the creator's is shown on the left (green name), whereas the most distant edition is on the right (red name).}
    \label{fig:errors-percent}
\end{figure}

Overall, we observe that all of our alignment strategies successfully reduce the original number of errors compared to the naive alignment. The \texttt{case} and \texttt{retokenize} strategies are the least effective. Meanwhile, the \texttt{mlm} and \texttt{propagate} strategies obtain better results. Meta-strategy \texttt{pipe} outperforms all of the singular strategies, confirming the interest of combining them.

\begin{figure*}[!htb]
    \includegraphics[width=\linewidth]{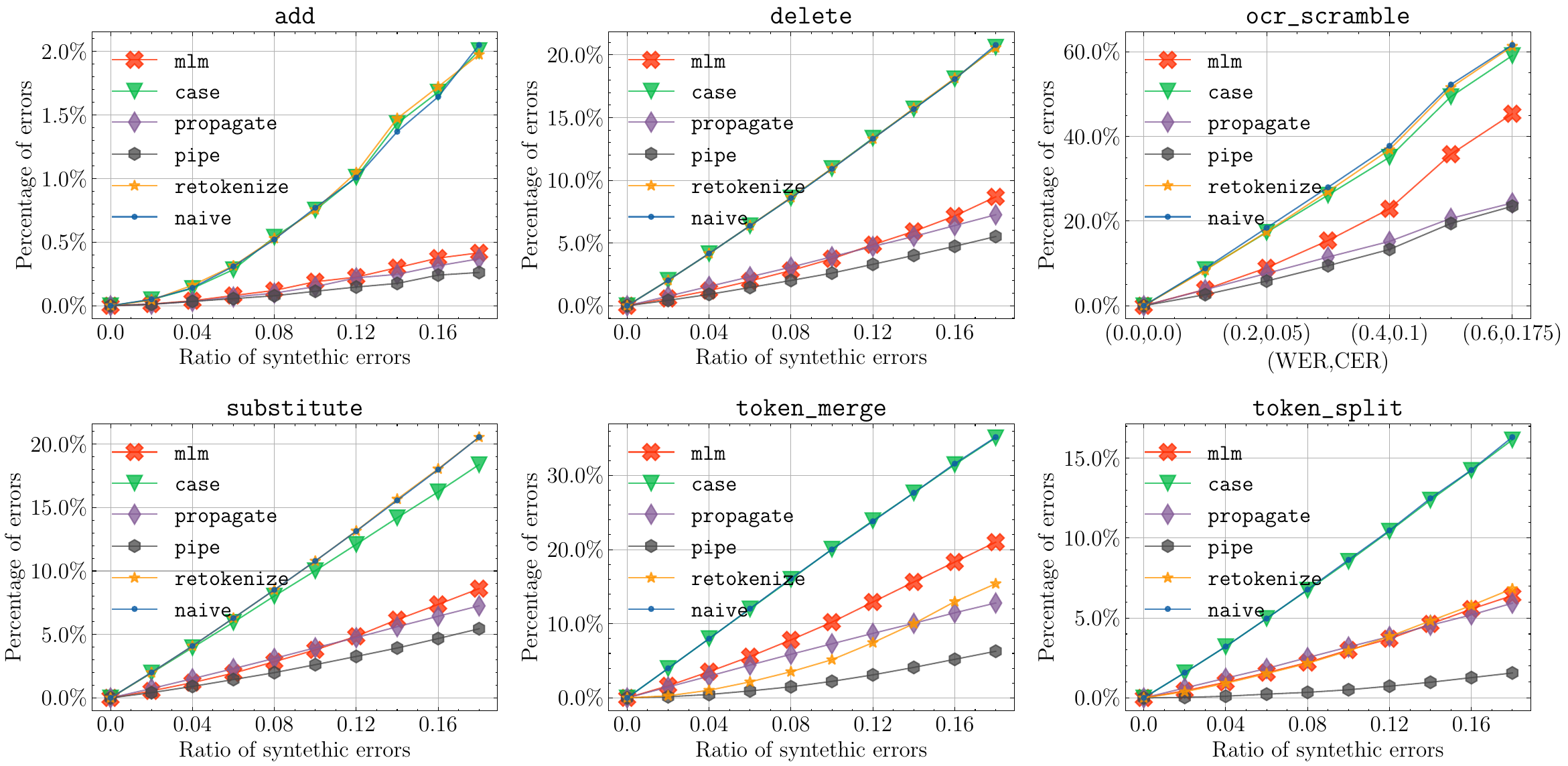}
    \caption{Percentage of alignment errors as a function of the ratio of synthetic errors added.}
    \label{fig:synthetic_errors}
\end{figure*}

As we hypothesized earlier, the alignment performance strongly depends on the proximity of the user's edition to the original text. Using the close editions and the best strategy results in a percentage of errors that does not exceed 1.3\%. Meanwhile, using the reworked 1831 edition of \textit{Frankenstein}, the censored UK edition of \textit{Moby Dick} or the modernized 1894 edition of \textit{Pride and Prejudice} yield substantially more errors. These results emphasize the need for the user to have a version of the data that is as close as possible to the original text.

Not all strategies are equal when it comes to runtime, as we show in Appendix~\ref{apdx:xp-edition-duration}. \texttt{mlm} and \texttt{pipe} in particular are costly, and can bring the alignment time up to the hour in the worst cases. Other strategies most often stay under a 10 seconds runtime.

\subsection{Synthetic Errors}
\label{sec:ResSyntheticErrors}

To better understand the effect of different possible degradations in the user's sequence and the impact of our additional alignment strategies, we add synthetic errors to the creator's edition of our corpus, creating a new synthetic user's edition, and to measure the impact it has on the performance of our alignment system. We experiment with six types of synthetic errors: 

\paragraph{\texttt{add}} We sample tokens from a dictionary, using their frequency in the considered novel, and add them to a uniformly sampled position in the text.

\paragraph{\texttt{substitute}} We sample tokens as in \texttt{add}, but replace them by others instead of adding new ones.

\paragraph{\texttt{delete}} We remove uniformly sampled tokens.

\paragraph{\texttt{tokens\_split}} We simulate tokenization errors, splitting uniformly sampled tokens.

\paragraph{\texttt{tokens\_merge}} We simulate tokenization errors by merging uniformly sampled consecutive tokens.

\paragraph{\texttt{ocr\_scramble}} We simulate realistic OCR issues using the \texttt{scrambledtext} library~\citep{Bourne2025-scrambledtext}.

\medskip

For all of these types of errors except \texttt{ocr\_scramble}, we produce $l \times r$ errors, with $l$ the length of the text and $r$ a ratio between 0 and 1. In practice, we consider error ratios from 0 to 0.2 for completeness, but we find large ratios unrealistic: counting additions, deletions and substitutions at the same time, we found the biggest ratio between novels of our corpus to be around 0.035. Regarding OCR errors, the \texttt{scrambledtext} library allows setting a target word error rate (WER) and a character error rate (CER). We survey the following (WER, CER) pairs: $\{(0, 0)$, $(0.1, 0.025)$, $(0.2, 0.05)$, $(0.3, 0.075)$, $(0.4, 0.10)$, $(0.5, 0.15)$, $(0.6, 0.175)\}$ following values presented by \citet{Bourne2025-scrambledtext} on the CA~\citep{Bourne2024}, SMH~\citep{Evershed2014-overproof} and BLN600~\citep{Booth2024} datasets. Note that OCR error levels beyond $(0.2, 0.05)$ correspond to heavy, difficult to recover levels of errors. We present examples of OCR error levels in Appendix~\ref{apdx:ocr-errors}.

Figure~\ref{fig:synthetic_errors} shows our results, leading to several observations. First, we are always able to fully align annotations when the user's text is identical to the creator's. We also note that the \texttt{pipe} meta-strategy outperforms other strategies in all cases, highlighting the interest of combining them. The \texttt{retokenize} strategy is very effective in case of token splitting or merging, but its performance is very weak in other settings, making it a specialized strategy for tokenization issues. \texttt{mlm} and \texttt{propagate} appear to be the best performing single strategies. Finally, we observe that adding new tokens with \texttt{add} errors only has a very weak impact on the number of errors compared to other types of errors. Heavy levels of OCR errors are difficult to recover from, but our method stays relatively successful even for moderate levels of errors (6.66\% of errors using the \texttt{pipe} strategy for a (WER, CER) pair of $(0.2, 0.05)$).

\section{Conclusion}

In this article, we presented a method to let a corpus creator legally share their annotations of copyrighted texts, provided the user of the corpus is in possession of this material. Our method supports more cases than covered by existing practices, such as situations where copyrighted works are not directly available online, preventing the user to access the exact same version of the data. Section~\ref{sec:ResSyntheticErrors} shows that our method is always able to fully align annotations when the user's content is identical to the creator's. Furthermore, our experiments in Section~\ref{sec:ResEditions} show that the alignment is successful even if the user owns a different (but sufficiently close) version of the material, as we reach a percentage of errors between 0.21\% and 1.3\% in that case. The number of errors, however, increases as the user's content diverges from the corpus creator's. This stems from our need to balance alignment performance and security, as an attacker with completely different data must not be able to access the corpus for our method to respect copyright. This also highlights the importance for the creator to provide to the user as much information as possible about the corpus (such as novel editions for literary text), to ease alignment.

As an illustration, we applied our alignment method to NER in Appendix~\ref{apdx:ner-results}. We expect that our sharing scheme can be applied for other types of tasks with token-level annotations such as POS tagging, coreference resolution, chunking or slot filling. Conceptually, annotations of non textual corpora may also be shared through adaptation of our technique, provided the target tasks can be formalized at the token level. We leave to future work the extension of our method to these different tasks and domains.

In this work, we limited ourselves to traditional tokenization schemess. Intuitively, we think there is potential for exploration, as certain types of errors may be easier to recover with different schemes. For example, OCR errors or misspellings may be easier to recover when using sub-word tokenization, as alignment errors would be limited to sub-words instead of entire words.

\section*{Limitations}

The severity of alignment errors is task-dependent: certain errors may be more important depending on the application context. We present additional results on NER in Appendix~\ref{apdx:ner-results}, but it is not feasible to study the impact of errors for all possible tasks.

Our alignment method could be adapted and applied to other sequential corpora. However, some of the additional alignment strategies we present are specific to natural language corpora and cannot be applied directly to other tasks. The \texttt{retokenize} strategy, for example, relies on the necessity to tokenize text for NLP tasks. The \texttt{case} strategy is limited to natural language. Masked language modeling could be applied to other tasks, but one needs an appropriately pretrained model to do so.

\section*{Acknowledgments}
We thank the reviewers and area chair for their valuable suggestions that strengthened the final version. 
This research was partially supported by the National Science and Technology Council (NSTC), Taiwan, under Grant No. 112-2221-E-001-016-MY3 and by Academia Sinica under Grants Nos. 236d-1120205 and 235g-1150000. 

The authors acknowledge the use of AI assistants, which were strictly used for literature search. The authors are responsible for all of the content included in the article.

\bibliography{Amalvy2026.bib}

@Article{Ahmed2026,
  author    = {Ahmed, A. and Cooper, A. F. and Koyejo, S. and Liang, P.},
  title     = {Extracting books from production language models},
  journal   = {arXiv},
  year      = {2026},
  volume    = {cs.CL},
  pages     = {2601.02671},
  url       = {https://arxiv.org/abs/2601.02671},
}

@InProceedings{Alrashid2023,
  author    = {Alrashid, T. and Gaizauskas, R.},
  title     = {ScANT: A Small Corpus of Scene-Annotated Narrative Texts},
  booktitle = {6th Workshop on Narrative Extraction From Texts},
  year      = {2023},
  series    = {CEUR Workshop Proceeedings},
  pages     = {143-149},
  url       = {https://www.di.ubi.pt/~jpaulo/T2S/paper15.pdf},
}

@InProceedings{Amalvy2024e,
  author    = {Amalvy, A. and Labatut, V. and Dufour, R.},
  title     = {The Role of Natural Language Processing Tasks in Automatic Literary Character Network Construction},
  booktitle = {31st International Conference on Computational Linguistics},
  year      = {2025},
  pages     = {8462-8473},
  url       = {https://aclanthology.org/2025.coling-main.566/},
}

@TechReport{Amalvy2024-novelties_ner,
    author      = {Amalvy, A. and Labatut, V.},
    title       = {Annotation Guidelines for Corpus Novelties: Part 1 –- Named Entity Recognition},
    institution = {Avignon Universit\'{e}},
    year        = {2024},
    url         = {https://hal.science/hal-04715338},
}

@InProceedings{Bamman2019-litbank,
  title        = {An annotated dataset of literary entities},
  author       = {Bamman, D. and Popat, S. and Shen, S.},
  booktitle    = {Conference of the North {A}merican Chapter of the Association for Computational Linguistics: Human Language Technologies},
  volume       = {1},
  year         = {2019},
  doi          = {10.18653/v1/N19-1220},
  pages        = {2138-2144},
}

@InProceedings{Batu2004-trace_reconstruction,
  author       = {Batu, T. and Kannan, S. and Khanna, S. and McGregor,
                  A.},
  title        = {Reconstructing strings from random traces},
  year         = {2004},
  booktitle    = {Fifteenth Annual ACM-SIAM Symposium on Discrete
                  Algorithms},
  pages        = {910–918},
  doi          = {10.5555/982792.982929},
}

@Article{Beelen2022,
  author    = {Beelen, K. and Lawrence, J. and Wilson, D. C. S. and Beavan, D.},
  title     = {Bias and representativeness in digitized newspaper collections: Introducing the environmental scan},
  journal   = {Digital Scholarship in the Humanities},
  year      = {2022},
  volume    = {38},
  number    = {1},
  pages     = {1-22},
  doi       = {10.1093/llc/fqac037},
}

@InProceedings{Booth2024,
	author = {Booth, C. W. and Thomas, A. and Gaizauskas, R.},
	title = {BLN600: A Parallel Corpus of Machine/Human Transcribed Nineteenth Century Newspaper Texts},
	year = 2024,
	booktitle = {2024 Joint International Conference on Computational Linguistics, Language Resources and Evaluation (LREC-COLING 2024)},
	series = {LREC-COLING},
	pages = {2440–2446},
	doi = {10.63317/525gz987px5s},
	collection = {LREC-COLING}
}

@InProceedings{Bost2020-serial_speakers,
  title        = {Serial Speakers: a Dataset of {TV} Series},
  author       = {Bost, X. and Labatut, V. and Linares, G.},
  booktitle    = {Twelfth Language Resources and Evaluation Conference},
  year         = {2020},
  url          = {https://aclanthology.org/2020.lrec-1.525/},
  pages        = {4256-4264},
}

@InProceedings{Bourgois2025,
  author    = {Bourgois, A. and Poibeau, T.},
  title     = {The Elephant in the Coreference Room: Resolving Coreference in Full-Length French Fiction Works},
  booktitle = {8th Workshop on Computational Models of Reference, Anaphora and Coreference},
  year      = {2025},
  url       = {https://arxiv.org/abs/2510.15594},
}

@Article{Bourne2024,
	author = {Bourne, J.},
	title = {{CLOCR-C: Context Leveraging OCR Correction with Pre-trained Language Models}},
	journal = {arXiv},
	year = 2024,
	volume = {cs.CL},
	pages = {2408.17428v2},
	url = {https://arxiv.org/abs/2408.17428v2}
}

@Article{Bourne2025-scrambledtext,
  author       = {Bourne, J.},
  year         = {2025},
  journal      = {International Journal on Document Analysis and Recognition},
  title        = {Scrambled text: fine-tuning language models for OCR error correction using synthetic data},
  doi          = {10.1007/s10032-025-00522-0},
}

@Article{Buick2024,
  author    = {Buick, A.},
  title     = {Copyright and AI training data--transparency to the rescue?},
  journal   = {Journal of Intellectual Property Law and Practice},
  year      = {2024},
  volume    = {20},
  number    = {3},
  pages     = {182-192},
  doi       = {10.1093/jiplp/jpae102},
}

@Article{Camp2012,
  author    = {van de Camp, M. and van den Bosch, A.},
  title     = {The socialist network},
  journal   = {Decision Support Systems},
  year      = {2012},
  volume    = {53},
  number    = {4},
  pages     = {761-769},
  doi       = {10.1016/j.dss.2012.05.031},
}

@InProceedings{Chen2012-dna_privacy,
  author       = {Chen, Y. and Peng, B. and Wang, X. and Tang, H.},
  title        = {Large-scale privacy-preserving mapping of human genomic sequences on hybrid clouds},
  booktitle    = {Network and Distributed System Security Symposium},
  year         = {2012},
  url          =
                  {https://www.ndss-symposium.org/ndss2012/ndss-2012-programme/large-scale-privacy-preserving-mapping-human-genomic-sequences-hybrid-clouds/},
}

@Article{Chun2025,
  author    = {Chun, Ye Eun and Hwang, Taeyoon and Hwang, Seung-won and Kim, Byung-Hak},
  title     = {{CREFT}: Sequential Multi-Agent LLM for Character Relation Extraction},
  journal   = {arXiv},
  year      = {2025},
  volume    = {cs.CL},
  pages     = {2505.24553},
  url       = {https://arxiv.org/abs/2505.24553},
}

@Article{Dekker2019-charnets,
  author       = {Dekker, N. and Kuhn, T. and van Erp, M.},
  journal      = {PeerJ Computer Science},
  title        = {Evaluating named entity recognition tools for
                  extracting social networks from novels},
  year         = {2019},
  pages        = {e189},
  volume       = {5},
  doi          = {10.7717/peerj-cs.189},
}

@InProceedings{Derczynski2017,
	title = {Results of the {WNUT}2017 Shared Task on Novel and Emerging Entity Recognition},
	author = {Derczynski, L. and Nichols, E. and van Erp, M. and Limsopatham, N.},
	booktitle = {3rd Workshop on Noisy User-generated Text},
	year = {2017},
	doi = {10.18653/v1/W17-4418},
	pages = {140-147}
}

@Misc{Evershed2014-overproof,
	url = {http://overproof.projectcomputing.com/evaluation},
	year = {2014},
	title = {Overproof - Evaluation},
	author = {Evershed, J. and Fitch, K.}
}

@InProceedings{Han2021b,
  author     = {Han, S. and Seo, S. and Kang, M. and Kim, J. and Choi, N. and Song, M. and Choi, J. D.},
  title      = {FantasyCoref: Coreference Resolution on Fantasy Literature Through Omniscient Writer’s Point of View},
  booktitle  = {4th Workshop on Computational Models of Reference, Anaphora and Coreference},
  year       = {2021},
  pages      = {24-35},
  doi        = {10.18653/v1/2021.crac-1.3},
  url        = {https://aclanthology.org/2021.crac-1.3/},
}

@Article{Henderson2023,
  author    = {Henderson, P. and Li, X. and Jurafsky, D. and Hashimoto, T. and Lemley, M. A. and Liang, P.},
  title     = {Foundation Models and Fair Use},
  journal   = {Journal of Machine Learning Research},
  year      = {2023},
  volume    = {24},
  number    = {400},
  pages     = {1-79},
  url       = {https://www.jmlr.org/papers/v24/23-0569.html},
}

@Article{Hussey2025,
	author = {Hussey, I.},
	title = {Data is not available upon request},
	journal = {Meta-Psychology},
	year = 2025,
	volume = 9,
	doi = {10.15626/mp.2023.4008}
}

@InProceedings{Johnson2024,
  author    = {Johnson, N. and Bertsch, A. and Strubell, E.},
  title     = {FicSim: An Ethically Constructed Dataset for Long-Context Semantic Similarity Comparison within FictionWorkshop on Creativity \& Generative AI},
  booktitle = {NeurIPS Workshop on Creativity \& Generative AI},
  year      = {2024},
  url       = {https://creativity-ai.github.io/assets/papers/21.pdf},
}

@InProceedings{Kang2016-dna_privacy,
  author       = {Kang, S. and Aung, K. M. M. and Veeravalli, B.},
  title        = {Towards Secure and Fast Mapping of Genomic Sequences
                  on Public Clouds},
  year         = {2016},
  doi          = {10.1145/2898445.2898448},
  booktitle    = {4th ACM International Workshop on Security in Cloud
                  Computing},
  pages        = {59–66},
  numpages     = {8},
}

@InProceedings{Lazaridou2021,
  author    = {Lazaridou, A. and Kuncoro, A. and Gribovskaya, E. and Agrawal, D. and Liška, A. and Terzi, T. and Gimenez, M. and de Masson d'Autume, C. and Kocisky, T. and Ruder, S. and Yogatama, D. and Cao, K. and Young, S. and Blunsom, P.},
  title     = {Mind the gap: assessing temporal generalization in neural language models},
  booktitle = {35th International Conference on Neural Information Processing Systems},
  year      = {2021},
  pages     = {29348-29363},
  url       = {https://proceedings.neurips.cc/paper_files/paper/2021/hash/f5bf0ba0a17ef18f9607774722f5698c-Abstract.html},
}

@Article{Levenshtein2001-seqa,
  author       = {Levenshtein, V.},
  journal      = {IEEE Transactions on Information Theory},
  title        = {Efficient reconstruction of sequences},
  year         = {2001},
  volume       = {47},
  number       = {1},
  pages        = {2-22},
  doi          = {10.1109/18.904499},
}

@Article{Levenshtein2001-seqb,
  author       = {Levenshtein, V.},
  title        = {Efficient Reconstruction of Sequences from Their
                  Subsequences or Supersequences},
  journal      = {Journal of Combinatorial Theory, Series A},
  volume       = {93},
  number       = {2},
  pages        = {310-332},
  year         = {2001},
  doi          = {https://doi.org/10.1006/jcta.2000.3081},
}

@Article{Lu2021-dna_privacy,
  author       = {Lu, D. and Zhang, Y. and Zhang, L. and Wang, H. and
                  Weng, W. and Li, L. and Cai, H.},
  title        = {Methods of privacy-preserving genomic sequencing
                  data alignments},
  journal      = {Briefings in Bioinformatics},
  volume       = {22},
  number       = {6},
  year         = {2021},
  doi          = {10.1093/bib/bbab151},
}

@Article{Melanie-Becquet2024,
  author    = {M\'{e}lanie-Becquet, F. and Barr\'{e}, J. and Seminck, O. and Plancq, C. and Naguib, M. and Pastor, M. and Poibeau, T.},
  title     = {{BookNLP-fr}, the {F}rench Versant of {BookNLP}. A Tailored Pipeline for 19th and 20th Century {F}rench Literature},
  journal   = {Journal of Computational Literary Studies},
  year      = {2024},
  volume    = {3},
  number    = {1},
  pages     = {1-34},
  doi       = {10.48694/JCLS.3924},
}

@Article{Mete2015-dna_privacy,
  title        = {Privacy preserving processing of genomic data: A
                  survey},
  journal      = {Journal of Biomedical Informatics},
  volume       = {56},
  pages        = {103-111},
  year         = {2015},
  doi          = {10.1016/j.jbi.2015.05.022},
  author       = {Mete, A. and A., O. and Bugra, O. and M., Ş.},
}

@InProceedings{Muzny2017-sieve_quote_attribution,
  author       = {Muzny, G. and Fang, M. and Chang, A. and Jurafsky,
                  D.},
  booktitle    = {15th Conference of the European Chapter of the
                  Association for Computational Linguistics},
  title        = {A Two-stage Sieve Approach for Quote Attribution},
  year         = {2017},
  pages        = {460-470},
  volume       = {1},
  url          = {https://aclanthology.org/E17-1044},
}

@InProceedings{Ramshaw1995-iob,
  title	     = {Text Chunking using Transformation-Based Learning},
  author       = {Ramshaw, L. and Marcus, M.},
  booktitle    = {Third Workshop on Very Large Corpora},
  year	     = {1995},
  url	         = {https://aclanthology.org/W95-0107},
}

@Article{Ratcliff1988-diff,
  title        = {Pattern Matching: The Gestalt Approach},
  journal      = {Dr. Dobb's Journal},
  volume       = {July 1988},
  pages        = {46},
  author       = {Ratcliff, J. W. and Metzener, D.},
  year         = {1988},
}

@InProceedings{Tjong2003,
	title = {Introduction to the {C}o{NLL}-2003 Shared Task: Language-Independent Named Entity Recognition},
	author = {Tjong Kim Sang, E. F. and De Meulder, F.},
	booktitle = {7th Conference on Natural Language Learning},
	year = {2003},
	url = {https://aclanthology.org/W03-0419},
	pages = {142-147}
}

@InProceedings{Vishnubhotla2022-PDNC,
  title        = {The Project Dialogism Novel Corpus: A Dataset for Quotation Attribution in Literary Texts},
  author       = {Vishnubhotla, K. and Hammond, A. and Hirst, G.},
  booktitle    = {13th Language Resources and Evaluation  Conference},
  year         = {2022},
  url          = {https://aclanthology.org/2022.lrec-1.628},
  pages        = {5838-5848},
}

@InProceedings{Warner2025-modernbert,
  title        = {Smarter, Better, Faster, Longer: A Modern
                  Bidirectional Encoder for Fast, Memory Efficient,
                  and Long Context Finetuning and Inference},
  author       = {Warner, B. and Chaffin, A. and Clavié, B. and
                  Weller, O. and Hallström, O. and Taghadouini, S. and
                  Gallagher, A. and Biswas, R. and Ladhak, F. and
                  Aarsen, T. and Adams, G. T. and Howard, J. and Poli,
                  I.},
  booktitle    = {63rd Annual Meeting of the Association for
                  Computational Linguistics},
  year         = {2025},
  doi          = {10.18653/v1/2025.acl-long.127},
  pages        = {2526-2547},
}

@Article{Wei2024,
  author       = {Wei, H. and Schwartz, M. and Ge, G.},
  title        = {Reconstruction From Noisy Substrings},
  year         = {2024},
  volume       = {70},
  number       = {11},
  doi          = {10.1109/TIT.2024.3454119},
  journal      = {IEEE Transactions on Information Theory},
  pages        = {7757–7776},
}

@Article{Wei2025-deepseek_ocr,
  title        = {DeepSeek-OCR: Contexts Optical Compression},
  author       = {Wei, H. and Sun, Y. and Li, Y.},
  year         = {2025},
  volume       = {cs.CV},
  url          = {https://arxiv.org/abs/2510.18234},
  pages        = {2510.18234},
  journal      = {arXiv},
}

@InProceedings{Zhao2025a,
  author     = {Zhao, H. and Yan, Y. and Zhu, S. and Liu, H. and Jia, Y. and Zan, H. and Peng, M.},
  title      = {GenWebNovel: A Genre-oriented Corpus of Entities in Chinese Web Novels},
  booktitle  = {31st International Conference on Computational Linguistics},
  year       = {2025},
  pages      = {3836-3849},
  url        = {https://aclanthology.org/2025.coling-main.259/},
}

\appendix

\section{Legal Aspects}
\label{sec:Legal}

\paragraph{Legal Framework.} The international baseline for the protection of copyrighted literary works is set by the \textit{Berne Convention for the Protection of Literary and Artistic Works}\footnote{\url{https://www.wipo.int/wipolex/en/text/283698}}. In particular, it forbids their unauthorized reproduction, in whole or in part (Article~9), and their communication to the public (Articles~11 \&~11bis). 

Under directive \textit{2001/29/EC}\footnote{\url{https://eur-lex.europa.eu/eli/dir/2001/29/oj/eng}}, European Union law forbids the direct or indirect unauthorized reproduction of copyrighted works or substantial parts thereof (Article~2), and their unauthorized communication to the public (Article~3). Furthermore, the Court of Justice of the European Union (CJEU) emphasizes that any use containing recognizable expression is prohibited (\textit{Infopaq C-5/08}\footnote{\url{https://ipcuria.eu/case?reference=C-5/08}}, \textit{Pelham C-476/17}~\footnote{\url{https://ipcuria.eu/case?reference=C-476/17}}). US copyright law (Copyright Act, Title 17 U.S.C.\footnote{\url{https://www.copyright.gov/title17/}} also forbids reproducing or copying these works (§106(1)) or distributing them (§106(3), §106(4)) without authorization. 

In the UK, the \textit{Copyright, Designs and Patents Act}\footnote{\url{https://www.legislation.gov.uk/ukpga/1988/48/contents}} (CDPA) similarly forbids copying or reproducing these works or substantial parts (ss. 16–17 CDPA), and issuing copies to the public (s.18 CDPA). Other common law countries and major jurisdictions apply the same rules. As a consequence, sharing the original literary text, even partially and even digitally, is in principle forbidden. Whether sharing \textit{excerpts} of literary works for research purpose (as done by~\cite{Dekker2019-charnets, Zhao2025a}) constitutes fair use is another debate, which we do not discuss here: we are only interested in sharing \textit{fully} annotated works.

\paragraph{Sharing Plain Annotations.} Our method does not involve sharing \textit{directly} any copyrighted material, but 1) a hashed version of the original text, and 2) the associated annotations authored by the researchers creating the corpus. These annotations are shared in plain text, but they are technical data, and not part of the original literary work. Put differently, they are distinct from the expressive content of the literary work (i.e. its actual words, narrative voice, the author's stylistic choices). As such, they are not protected by the laws mentioned before. 

In the EU, directive \textit{(EU) 2019/790}\footnote{\url{https://eur-lex.europa.eu/eli/dir/2019/790/oj/eng}} states that analytical outputs and metadata are legally distinct from the protected works themselves, and explicitly permits research mining for non-expressive results (Articles~3 \&~4), even of copyrighted works, provided outputs are non-substitutive (i.e. the original text cannot be recovered based on these data). In the USA, courts recognize that uses which do not communicate the expressive content and are transformative or functional can be fair use (17 U.S.C. §107). For instance, case \textit{Authors Guild v. Google}\footnote{\url{https://www.copyright.gov/fair-use/summaries/authorsguild-google-2dcir2015.pdf}}, 804 F.3d 202 (2d Cir. 2015) concluded that Google Books’ scanning for indexing and search was illustrative of non-expressive fair use. 

In the UK, non-expressive computational uses of lawfully accessed works are permitted (ss.29A, 29B CDPA), including sharing outputs that do not contain readable or recognizable parts of the work. Other major jurisdictions implement similar rules regarding the technical data extracted from literary text.

\paragraph{Sharing Hashed Tokens.} Let us now focus on the hashed tokens of the copyrighted text. The original text is never shared, reproduced, or communicated as part of the corpus. Instead, each token of the source text is transformed using a non-reversible cryptographic hashing function. Only these hashed identifiers (together with the linguistic annotations produced by the researchers), are disseminated. The resulting corpus contains no readable text, no identifiable excerpts, and no information allowing access to or reconstruction of the original works. The corpus is unusable without prior access to the original text: the user must independently possess the work, and has to locally apply the same hashing procedure in order to align the annotations with their own plain text. As a consequence, there is no reproduction of the protected text, no communication of the original work to the public, and no creation of derivative works. 

In the Berne Convention, reproduction (Article~9), communication (Articles~11 \&~11bis), and adaptation (Article~12) only apply to recognizable expression: token hashes do not convey the expression of the novel, they are non-recognizable technical identifiers. The same reasoning applies to EU law, as without possession of the original work, the hashes are useless. CJEU cases (Infopaq C-5/08, Pelham C-476/17) require that identifiable expression is reproduced for infringement: hashes do not meet this criterion. US and UK laws similarly do not consider hashes as reproductions or distribution of expressive content. 

\paragraph{Reliability of the Hashing.} Based on these observations, a question arises: what about the \textit{reliability} of the hashing scheme? International copyright law does not require absolute, information‑theoretic irreversibility, but rather focuses on whether the material shared by the corpus' creator objectively communicates expressive content or makes it reasonably accessible to the public. If reversing the hashes would require disproportionate technical effort, the corpus would still be treated as non‑expressive under all major jurisdictions. As a consequence, there is no statutory minimum security level to be implemented in our hashing scheme, but there is a standard of practical non-reconstructability. This is the reason why our method lets the creator control the length of the hashes, which directly impacts the vulnerability of the hashing scheme to attacks.

The strategies that we propose in Section~\ref{sec:Strat} to improve the robustness of our method against misalignment issues weaken this argument, though. Indeed, they could be used to facilitate the reconstruction of the original text without access to this text, at least in theory. However, we must stress that this is not feasible in practice, as these strategies are just ancillary mechanisms that require the user to have access to a text extremely similar to the original content. Strategy \texttt{propagate} has the strongest effect on misaligned hashes (cf. Figure~\ref{fig:errors-percent}), but it is limited to tokens that are already known by the user. Strategies \texttt{retokenize} and \texttt{case} have a very marginal effect, and require the user to have access to the original text, even if incorrectly tokenized or capitalized. Strategy \texttt{mlm} has a slightly stronger effect, but it provides no guarantee that the MLM-generated token is the same as in the original text. Moreover, it is efficient only if the textual context of the missing token is known by the user, which therefore still has to prove they have access to the original material. In practice, as shown by our experimental results, even when dealing with two distinct editions of the same novel, we obtain up to 8 \% incorrect tokens (cf. Section~\ref{sec:ResEditions}). Applying our method from scratch, without possessing a very similar version of the original text, leads to some content very different from the targeted literary work. In conclusion, we think that our method minimizes legal risk related to copyright infringement, and is aligned with best practices for responsible data sharing in natural language processing research.

\section{Corpus Details}
\label{apdx:data-details}

Table~\ref{tab:data-sources} indicates where we obtained each novel edition used in our experiments. For all editions of \textit{Frankenstein}, we obtain the text through the \href{https://frankensteinvariorum.org/}{Frankenstein Variorum} website. 

In the case of \textit{Moby Dick}, we obtain the text of the U.S. edition through \href{https://en.wikisource.org/wiki/Moby-Dick_(1851)_US_edition}{Wikisource}. Since we were unable to find a digital edition of the original U.K. edition, we use Deepseek-OCR~\citep{Wei2025-deepseek_ocr} to extract text from the book images hosted at the \href{https://github.com/performant-software/mel-website/tree/master/images}{Melville Electronic Library}. 

For \textit{Pride and Prejudice}, we use Wikisource for both the \href{https://en.wikisource.org/wiki/Pride_and_Prejudice_(1813)}{first} (\texttt{PP-1813}) and \href{https://en.wikisource.org/wiki/Pride_and_Prejudice_(1817)}{second} (\texttt{PP-1817}) editions in our experiments. We include the \texttt{PP-1894} edition through \href{https://www.gutenberg.org/ebooks/1342}{project Gutenberg}. Since this version is illustrated, the raw project Gutenberg text contain descriptions of the included illustration: we manually remove these as a preprocessing step.

\begin{table*}[htb]
    \centering
    \begin{tabular}{|c|c|p{10cm}|}
        \hline
         \textbf{Novel Edition} & \textbf{Source Type} &  \textbf{Source Identifier} \\
         \hline
         \texttt{F-1818} & URL & \url{https://github.com/FrankensteinVariorum/fv-data/blob/master/preliminary-edition-data/1818_full_prelim.xml} \\
         \texttt{F-1823} & URL & \url{https://github.com/FrankensteinVariorum/fv-data/blob/master/preliminary-edition-data/1823_full_prelim.xml} \\
         \texttt{F-1831} & URL & \url{https://github.com/FrankensteinVariorum/fv-data/blob/master/preliminary-edition-data/1831_full_prelim.xml} \\
         \hline
         \texttt{MD-1851-US} & URL & \url{https://en.wikisource.org/wiki/Moby-Dick_(1851)_US_edition} \\
         \texttt{MD-1851-UK} & URL & \url{https://github.com/performant-software/mel-website/tree/master/images/md-british-v1} \newline \url{https://github.com/performant-software/mel-website/tree/master/images/md-british-v2} \newline \url{https://github.com/performant-software/mel-website/tree/master/images/md-british-v3} \\
         \texttt{MD-1988} & ISBN & \texttt{9780810102699} \\
         \hline
         \texttt{PP-1813} & URL & \url{https://en.wikisource.org/wiki/Pride_and_Prejudice_(1813)} \\
         \texttt{PP-1817} & URL & \url{https://en.wikisource.org/wiki/Pride_and_Prejudice_(1817)} \\
         \texttt{PP-1894} & URL & \url{https://www.gutenberg.org/ebooks/1342} \\
        \hline
    \end{tabular}
    \caption{Source from which we obtained each edition of our novels (\textbf{F}rankenstein, \textbf{M}oby \textbf{D}ick, \textbf{P}ride and \textbf{P}rejudice).}
    \label{tab:data-sources}
\end{table*}

\section{Alignment Runtime}
\label{apdx:xp-edition-duration}

\begin{figure}[htb]
    \centering
    \includegraphics[width=\linewidth]{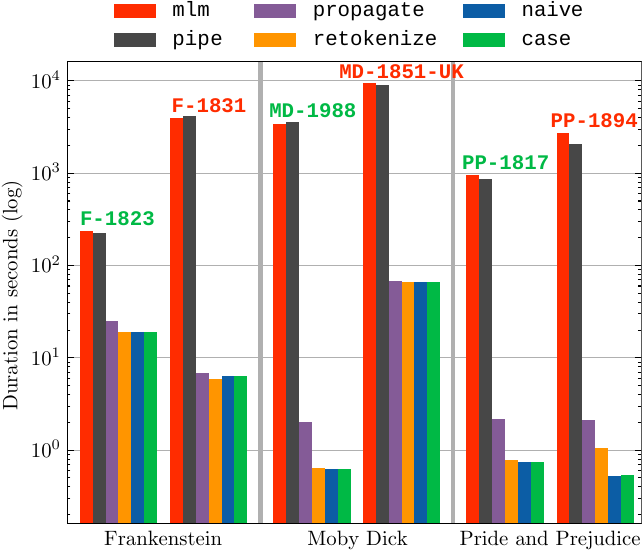}
    \caption{Duration of alignment depending on the strategy and user edition, in seconds.} 
    \label{fig:duration}
\end{figure}

In this section, we present more details on the runtime of alignment in our Section~\ref{sec:ResEditions} inter-edition experiments. As can be seen in Figure~\ref{fig:duration}, the \texttt{mlm} strategy is largely the most expensive, with a runtime that can go close to 3 hours for the very distant UK edition of \textit{Moby Dick}. The pipe strategy is close to the runtime of \texttt{mlm}, since it includes it. Meanwhile, other strategies do not break the 10 seconds barrier, except for the UK edition of \textit{Moby Dick} where the runtime is comprised between 1 and 2 minutes.

\section{Synthetic OCR Errors}
\label{apdx:ocr-errors}

During our experiments with synthetic errors, we survey different levels of OCR errors by experimenting with the following (WER, CER) pairs: $\{(0, 0)$, $(0.1, 0.025)$, $(0.2, 0.05)$, $(0.3, 0.075)$, $(0.4, 0.10)$, $(0.5, 0.15)$, $(0.6, 0.175)\}$. We present an example of applying these different levels of OCR errors in Table~\ref{tab:ocr_errors}. We empirically observe that errors beyond $(0.2, 0.05)$ (WER, CER) correspond to heavy OCR errors that are difficult to recover from.

\begin{table*}
    \centering
    \begin{tabular}{|c|p{11cm}|}
        \hline
        \textbf{Target (WER, CER)} & \textbf{Text} \\
        \hline
        (0.0, 0.0) & Our methods may seem strange and indirect. Even incomprehensible. But I assure you we know what we're doing. \\
        \hline
        (0.1, 0.025) & Our methods may seem strange an hindirect. Even incomprehensible. But I assure you weknow what we’re doing. \\
        \hline
        (0.2, 0.05) & Our methods may seem strange and indirect. Even incomprdhensible.i But I assure you we know what we’re doing. \\
        \hline
        (0.3, 0.075) & ODurmethods may seem strange and indirect. Even incomprehensible. But 4 assure you weknow whadr we’re dleing.. \\
        \hline
        (0.4, 0.1) & Our methods sylseem -tsangs 'end indirect. E,ven'  incomprehensible. But n assure yeu we now-what we’re doing. \\
        \hline
        (0.5, 0.15) & Our methlods  mna.yseem stbreoe anid indirect. Evei dn\'omtorohentsiblo. But I assure you weknow what we’re"" doing. \\
        \hline
        (0.6, 0.175) & oure mleottodls may seem str. .  ngo and idErect,., Evenincomprehensible. But  )I dassure o.n we know vla we’re dloing. \\
        \hline
    \end{tabular}
    \caption{Different levels of OCR errors applied to an example text from Philip K. Dick \textit{``Adjustment Team''}.}
    \label{tab:ocr_errors}
\end{table*}

\section{Results on Other Domains}
\label{apdx:results-other-domains}

To ensure that our technique is applicable to domains other than literary texts, we perform additional experiments on known NLP corpora: CoNLL 2003~\citep{Tjong2003} for the newswire domain and WNUT 2017~\citep{Derczynski2017} for the web domain. Since different versions of these datasets do not exist (as is the case with different editions of novels), we retorts to creating degraded versions of documents via synthetic errors and observing their effect on alignment performance as in Section~\ref{sec:ResSyntheticErrors}.

\begin{figure*}
    \centering
    \includegraphics[width=\linewidth]{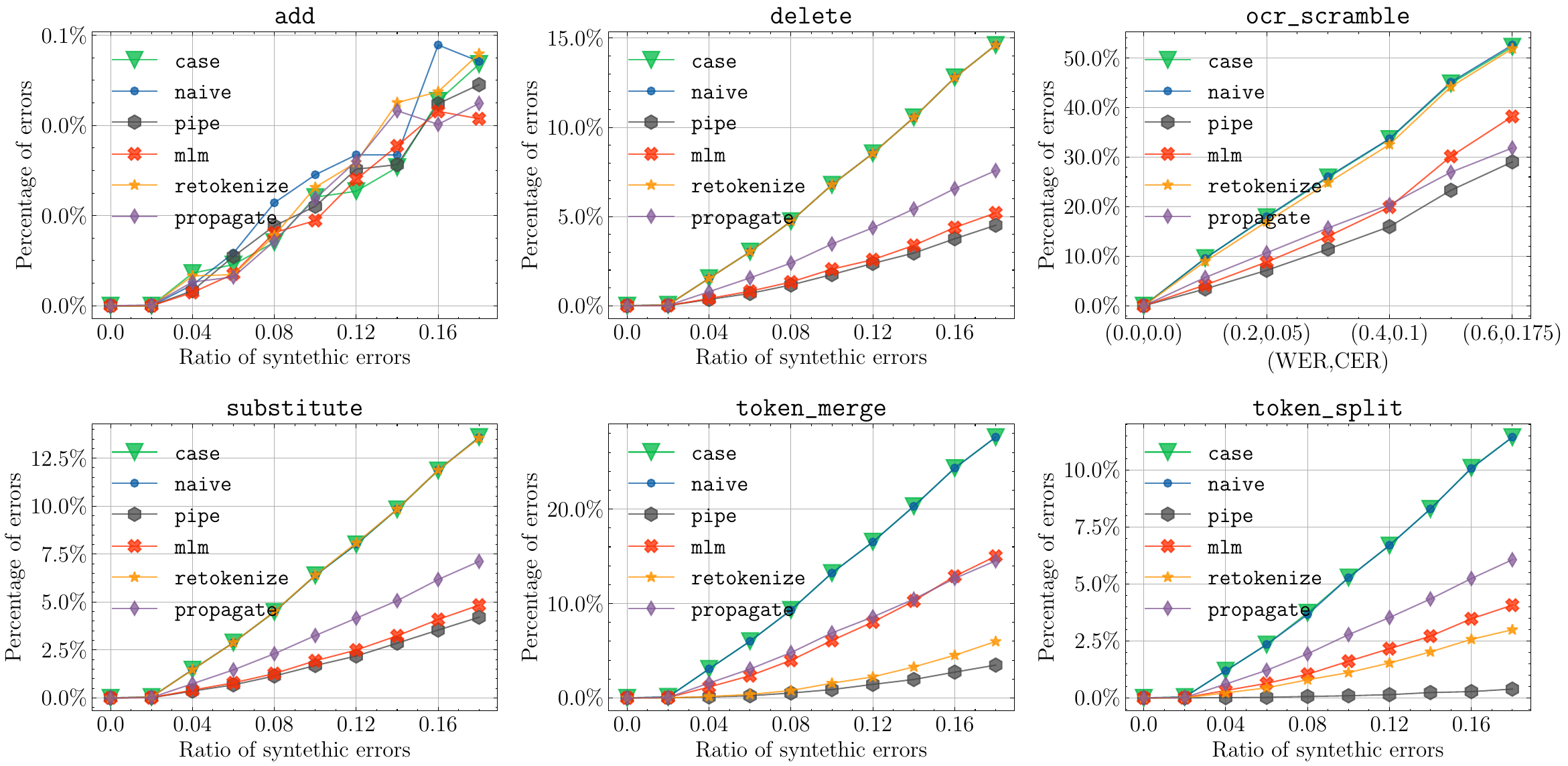}
    \caption{Percentage of alignment errors as a function of the ratio of synthetic errors added to the CoNLL 2003 dataset.}
    \label{fig:synthetic_errors_conll2003}
\end{figure*}

\begin{figure*}
    \centering
    \includegraphics[width=\linewidth]{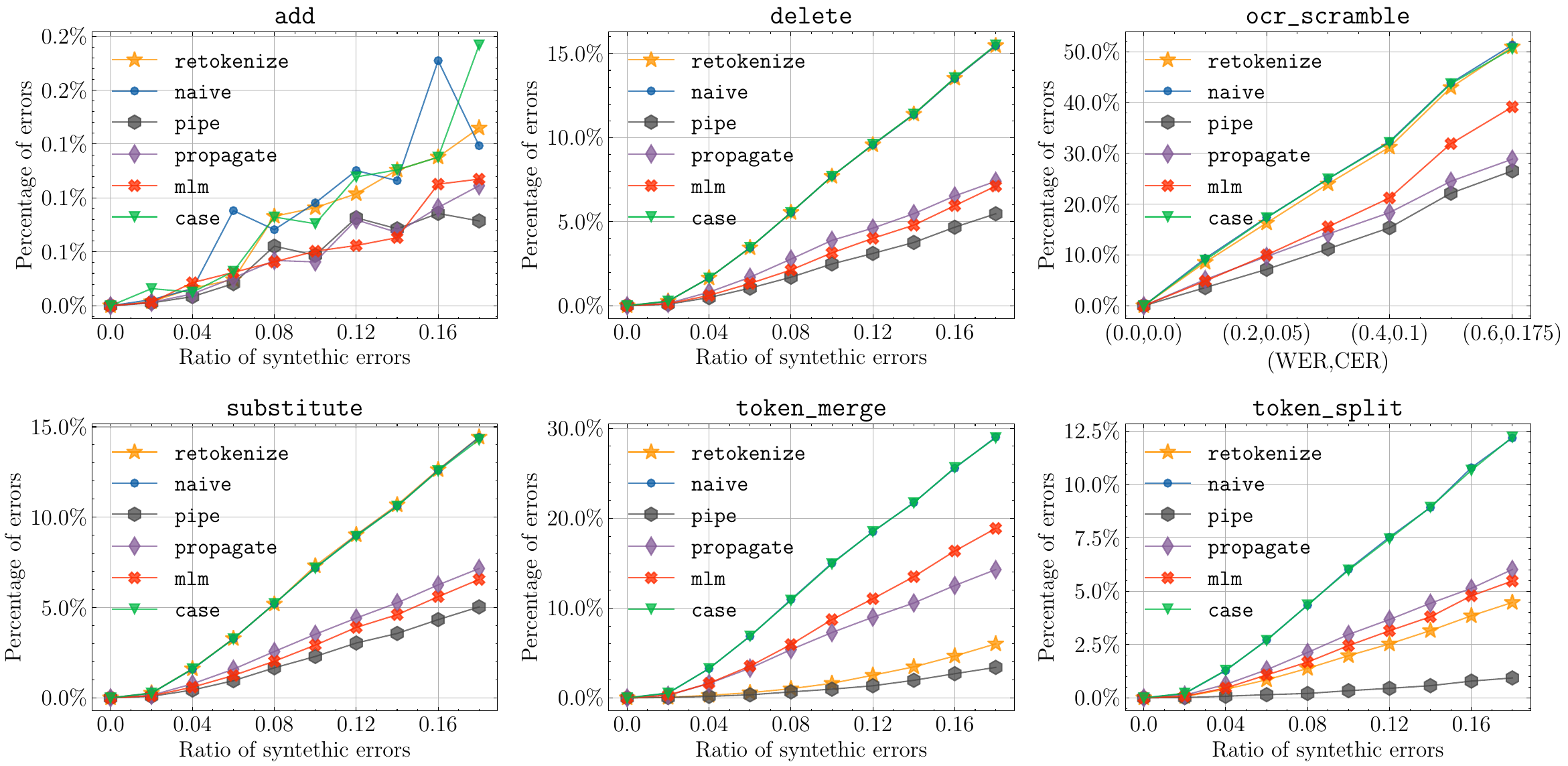}
    \caption{Percentage of alignment errors as a function of the ratio of synthetic errors added to the WNUT 2017 dataset.}
    \label{fig:synthetic_errors_wnut2017}
\end{figure*}

We plot the results of these experiments in Figure~\ref{fig:synthetic_errors_conll2003} (CoNLL 2003) and Figure~\ref{fig:synthetic_errors_wnut2017} (WNUT 2017).
For both of these datasets, we observe lower errors across the boards. This is at least partially due to the facts that these datasets are divided into examples that are smaller than chapters, facilitating alignment. The relative comparison between strategies yield similar results than on our corpus of novels, showing their mechanisms are not affected by text domain.

\section{Task-Specific Alignment Results}
\label{apdx:ner-results}

In the main text, we present results as a number of alignment errors. However, the severity of errors is task-dependent, as certain tokens may be more important as others. While it is not realistic to explore the impact of errors on every possible task, in this section we present additional results on NER.

To estimate the impact of alignment errors on NER, we use the NER-annotated version of \textit{Moby Dick} from the Novelties corpus~\citep{Amalvy2024-novelties_ner} as the source version. We use the \texttt{MD-1988} edition of \textit{Moby Dick}, as it is the closest from the Novelties version. We use a strict definition of the notion of error: if a single token from an entity was not aligned, we consider the entire entity as non-aligned. We obtain a percentage of errors of 0.82\% with our best strategy \texttt{pipe}, indicating that we are able to recover most of the text. However, we note a percentage of errors on entities of 3.52\%. While this is partially due to our strict definition of the notion of errors (a more lenient definition where the entire entity must be lost to be considered an error yields 2.93\% of errors), it also highlights that entity tokens are harder to align.

\begin{figure*}
    \centering
    \includegraphics[width=1.0\linewidth]{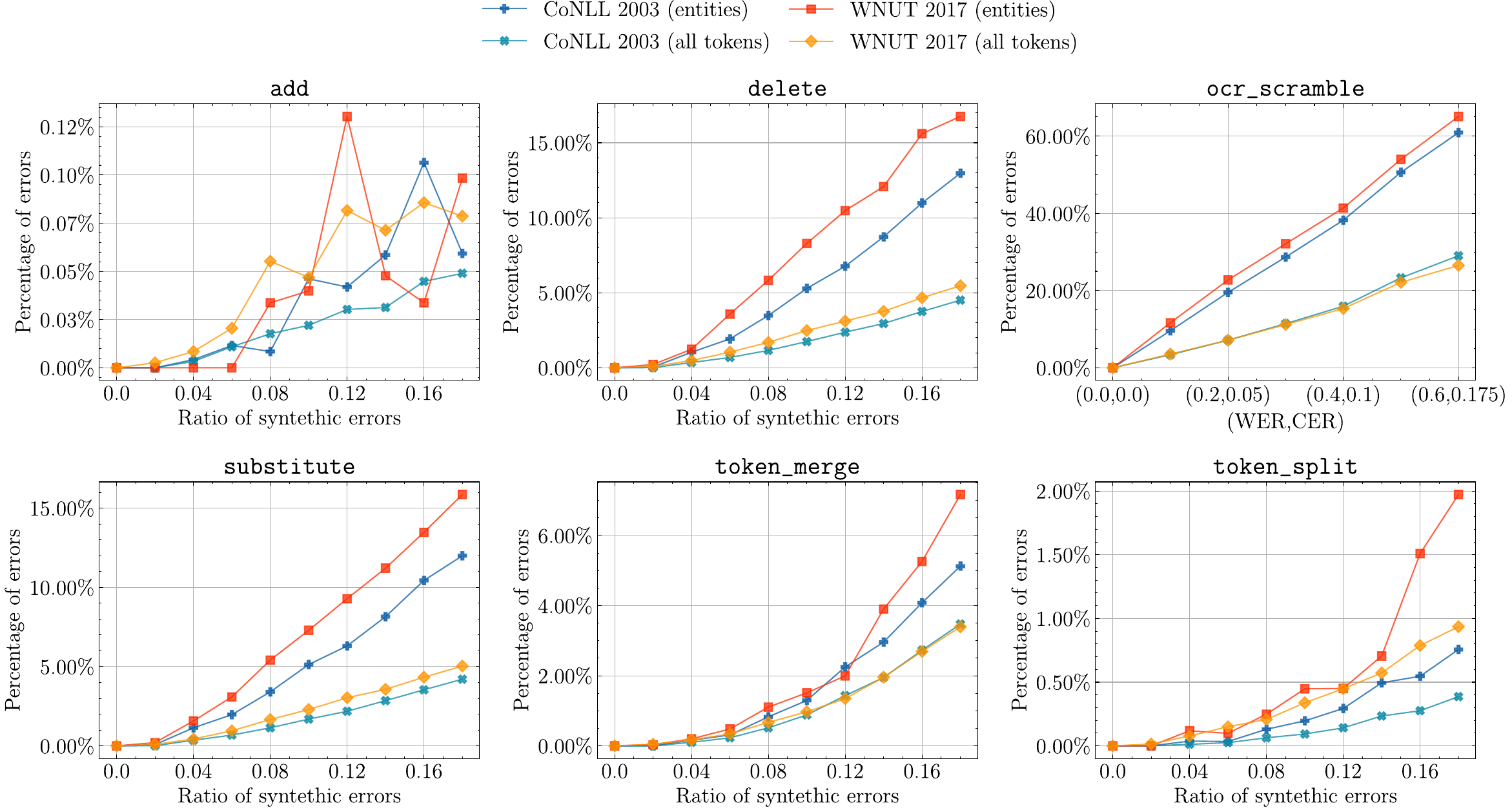}
    \caption{Percentage of alignment errors as a function of synthetic errors added to the CoNLL 2003 and WNUT 2017 datasets. We differentiate between the percentage of errors on \textit{all tokens} and on \textit{entities}. Only the \texttt{pipe} strategy is represented.}
    \label{fig:synthetic_errors_task_specific}
\end{figure*}

In order to extend our analysis to other domains, we also present results on the CoNLL 2003 (news) and WNUT 2017 (web) NER datasets in Figure~\ref{fig:synthetic_errors_task_specific}. For both of these datasets, we see that entities are generally more difficult to align than regular tokens. While Figure~\ref{fig:synthetic_errors_task_specific} shows our strict metric, we also generally note that behaviour with its more lenient version. We notice the biggest differences with the \texttt{ocr\_scramble}, \texttt{delete} and \texttt{substitute} types of errors.

\section{Optimal Parameters of Alignment Strategies}

\subsection{Masked Language Modeling Context Window Size}
\label{apdx:mlm-window}

In this section, we study the impact of the context window size on the performance of \texttt{mlm}, our masked language modeling alignment strategy (cf. Section~\ref{sec:Strat}).

\begin{figure}[htb]
    \centering
    \includegraphics[width=1.0\linewidth]{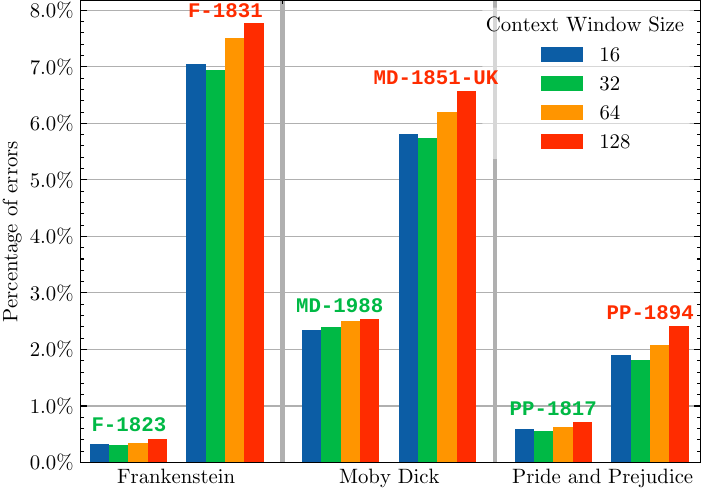}
    \caption{Percentage of errors using our \texttt{mlm} alignment strategy, depending on the context window size and user edition.} 
    \label{fig:mlm_params}
\end{figure}

Figure~\ref{fig:mlm_params} shows the performance of our \texttt{mlm} alignment strategy on our editions experiments from Section~\ref{sec:ResEditions}. We observe that a window size of 32 tokens generally better results for all but one edition. Beyond that, increasing window size seem to monotonically increase the number of errors.

\subsection{Order of Strategies in the \texttt{pipe} Meta-strategy}
\label{apdx:pipe-order}

\begin{figure}[htb]
    \centering
    \includegraphics[width=1.0\linewidth]{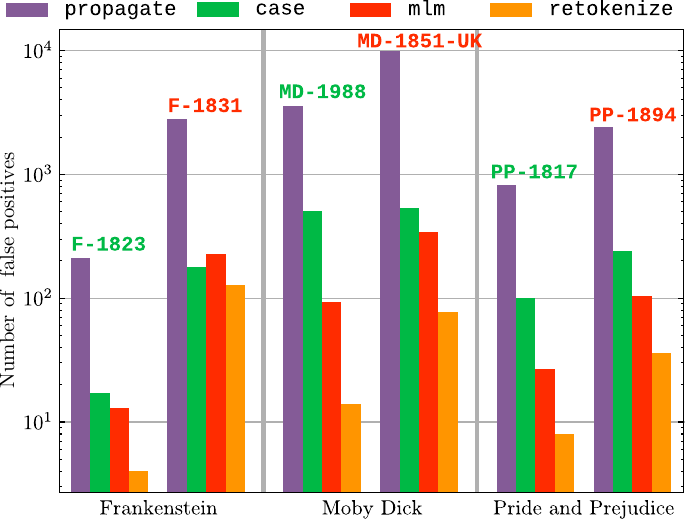}
    \caption{Number of false positives using different strategies \texttt{pipe}, depending on user edition.} 
    \label{fig:xp-edition-precision}
\end{figure}

The position of each strategy in the \texttt{pipe} meta-strategy influences performance. When a strategy corrects an alignment error, it will not be corrected further by the following strategies. Therefore, it is advantageous to place high precision strategies first in the pipeline. In the main text, we only report results with the best order we found for the \texttt{pipe} strategy. Figure~\ref{fig:xp-edition-precision} shows false positives recorded during our edition experiments of Section~\ref{sec:ResEditions} for each strategy. Interestingly, even though it is one of the best performing strategy, we notice that the propagate strategy has the lowest precision by far. By decreasing order of precision, we determine that the best ordering for the \texttt{pipe} strategy is \texttt{retokenize}, \texttt{mlm}, \texttt{case}, \texttt{propagate}.

\end{document}